\documentclass[10pt,twocolumn,letterpaper]{article}

\usepackage{cvpr}
\usepackage{times}
\usepackage{epsfig}
\usepackage{graphicx}
\usepackage{amsmath}
\usepackage{amssymb}
\usepackage{caption}
\usepackage{subcaption}
\usepackage{multirow}
\usepackage{floatrow}
\usepackage{color}

\usepackage{bbm}
\usepackage{array, tabularx}
\usepackage{tabulary}
\usepackage{wrapfig}
\usepackage{appendix}
\usepackage[pagebackref=true,breaklinks=true,letterpaper=true,colorlinks,bookmarks=false]{hyperref}

\cvprfinalcopy % *** Uncomment this line for the final submission

\def\cvprPaperID{1712} % *** Enter the CVPR Paper ID here

\ifcvprfinal\fi
\begin{document}

%%%%%%%%% TITLE
\title{End-to-end Learning of Action Detection from Frame Glimpses in Videos}

% \author{Serena Yeung\\
% Stanford University\\
% {\tt\small serena@cs.stanford.edu}
% % For a paper whose authors are all at the same institution,
% % omit the following lines up until the closing ``}''.
% % Additional authors and addresses can be added with ``\and'',
% % just like the second author.
% % To save space, use either the email address or home page, not both
% \and
% Olga Russakovsky\\
% Carnegie Mellon University\\
% {\tt\small olgarus@cmu.edu}
% \and
% Greg Mori\\
% Simon Fraser University\\
% {\tt\small mori@cs.sfu.ca}
% \and
% Li Fei-Fei\\
% Stanford University\\
% {\tt\small feifeili@cs.stanford.edu}
% }

\author{Serena Yeung\textsuperscript{1}, Olga Russakovsky\textsuperscript{1,2}, Greg Mori\textsuperscript{3}, Li Fei-Fei\textsuperscript{1}\\\\
\textsuperscript{1}Stanford University, \textsuperscript{2}Carnegie Mellon University, \textsuperscript{3}Simon Fraser University\\
{\tt\small serena@cs.stanford.edu, olgarus@cmu.edu, mori@cs.sfu.ca, feifeili@cs.stanford.edu}
}

\maketitle
\vspace{-0.1in}
%%%%%%%%% ABSTRACT
\begin{abstract}
In this work we introduce a fully end-to-end approach for action detection in videos that learns to directly predict the temporal bounds of actions. Our intuition is that the process of detecting actions is naturally one of observation and refinement: observing moments in video, and refining hypotheses about when an action is occurring. Based on this insight, we formulate our model as a recurrent neural network-based agent that interacts with a video over time. The agent observes video frames and decides both where to look next and when to emit a prediction. Since backpropagation is not adequate in this non-differentiable setting, we use REINFORCE to learn the agent's decision policy. Our model achieves state-of-the-art results on the THUMOS'14 and ActivityNet datasets while observing only a fraction (2\% or less) of the video frames.
\end{abstract}

\vspace{-0.15in}
%%%%%%%%% BODY TEXT
\section{Introduction}

Action detection in long, real-world videos is a challenging problem in computer vision. Algorithms must reason not only about whether an action occurs somewhere in a video, but also on the temporal extent of \textit{when} it occurs.  Most existing work~\cite{oneata2014lear, wangaction, karamanfast, yuanadsc} take the approach of building frame-level classifiers, running them exhaustively over a video at multiple temporal scales, and applying post-processing such as duration priors and non-maximum suppression. However, this indirect modeling of action localization is unsatisfying in terms of both accuracy as well as computational efficiency.

\begin{figure}[t]
\begin{center}
   \includegraphics[width=\linewidth]{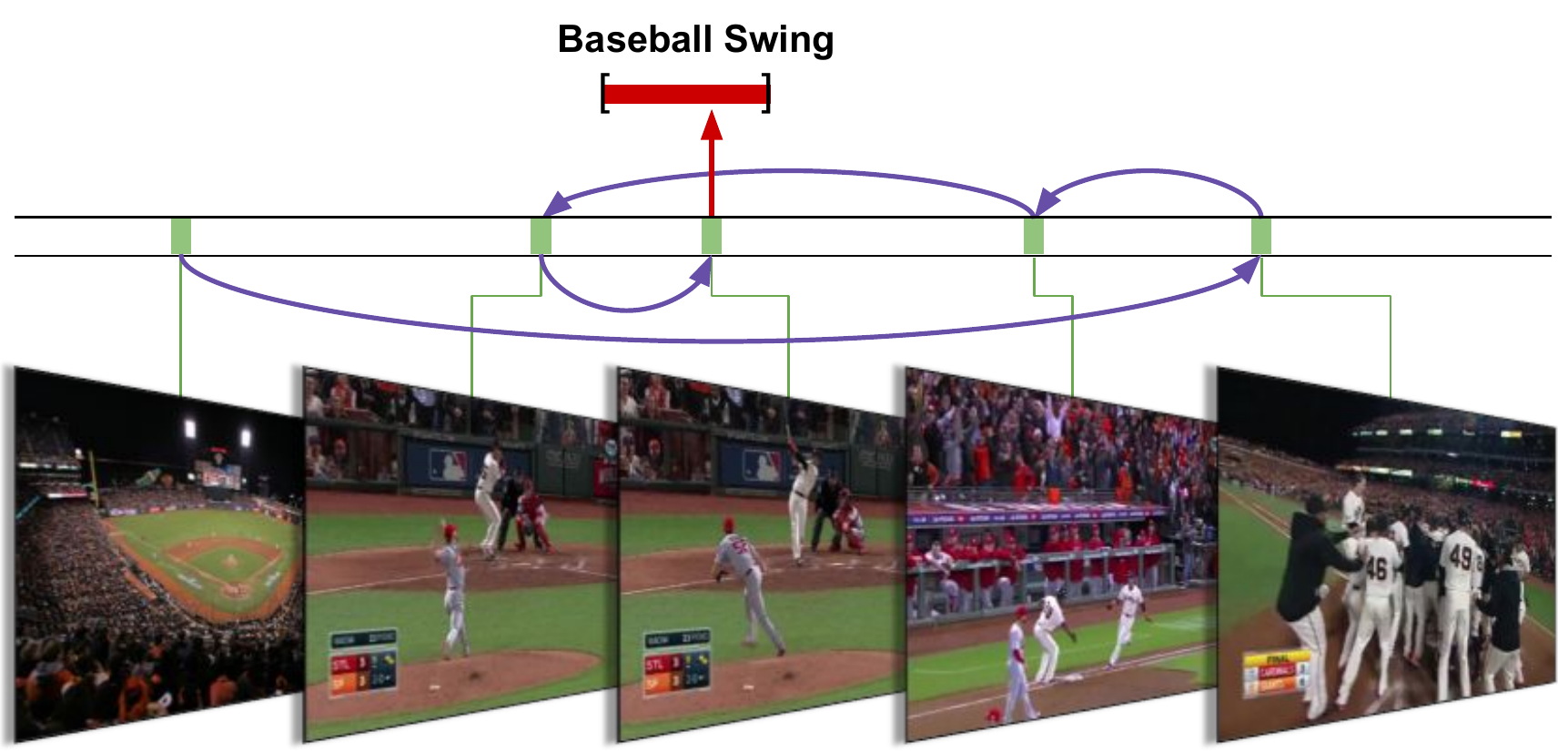}
\end{center}
\vspace{-0.15in}
   \caption{\small Action detection is a process of observation and refinement. Effectively choosing a sequence of frame observations allows us to quickly narrow down when the baseball swing occurs.}
\label{fig:pull}
\end{figure}

In this work, we introduce an end-to-end approach to action detection that reasons directly on the temporal bounds of actions. Our key intuition (Fig.~\ref{fig:pull}) is that the process of detecting an action is one of continuous, iterative observation and refinement. Given a single or a few frame observations, a human can already formulate hypotheses about when an action may occur.  We can then skip ahead or back some frames to verify, and quickly narrow down the action location (e.g. swinging a baseball bat in Fig.~\ref{fig:pull}). We are able to sequentially decide where to look and how to refine our hypotheses to obtain precise localization of the action with far less exhaustive search compared to existing algorithms.

Based on this intuition, we present \emph{a single coherent model} that takes a long video as input, and outputs the temporal bounds of detected action instances. Our model is formulated as an agent that learns a policy for sequentially forming and refining hypotheses about action instances. Casting this into a recurrent neural network-based architecture, we train the model in a fully end-to-end fashion using a combination of backpropagation and REINFORCE~\cite{williams1992simple}.

Our model draws inspiration from works that have used REINFORCE to learn spatial glimpse policies for image classification and captioning~\cite{mnih2014recurrent, ba2014multiple, sermanet2014attention, xu2015show}. However, action detection presents the additional challenge of how to handle a variable-sized set of structured detection outputs. \emph{To address this, we present a model that decides both which frame to observe next as well as when to emit a prediction, and we introduce a reward mechanism that enables learning this policy.} To the best of our knowledge, this is the first end-to-end approach for learning to detect actions in video.

We show that our model is able to reason effectively on the temporal bounds of actions, and achieve state-of-the-art performance on the THUMOS'14~\cite{THUMOS14} and ActivityNet~\cite{caba2015activitynet} datasets.  Moreover, because it learns policies for which frames to observe, or temporally glimpse, it is able to do so while observing only a fraction (2\% or less) of the frames.

\section{Related Work}
There is a long history of work in video analysis and activity recognition~\cite{moore2002recognizing, zhong2004detecting, blank2005actions, shi2006learning, laptev2008learning, gupta2009understanding, jhuang2013towards, kantorov2014efficient, zhu2015aligning}. For a survey we refer to Poppe~\cite{Poppe10} and Weinland et al.~\cite{WeinlandRB10}. Here we review recent work relevant to temporal action detection.

\noindent \textbf{Temporal action detection.} Canonical work in this vein is Ke et al.~\cite{ke07_iccv}. Rohrbach et al.~\cite{rohrbach2012database} and Ni et al.~\cite{ni2014multiple} use hand-centric and object-centric features, respectively, to detect fine-grained cooking actions in a fixed-camera kitchen setting. More related to our work is the unconstrained and untrimmed setting of the THUMOS'14 action detection dataset. Oneata et al.~\cite{oneata2014lear}, Wang et al.~\cite{wangaction}, Karaman et al.~\cite{karamanfast}, and Yuan et al.~\cite{yuanadsc} use fusions of dense trajectories, frame-level CNN features, and/or sound features in a sliding window framework to perform temporal action detection. Sun et al.~\cite{sun2015temporal} uses web images as a prior to improve detection performance.  Pirsiavash and Ramanan~\cite{PirsiavashR14} build grammars over complex actions and additionally detect sub-components in time.

Methods for spatio-temporal action detection have also been developed.  Within the context of ``unconstrained'' internet videos, this includes a body of work on spatio-temporal action proposals~\cite{yao2010hough, lan2011discriminative, tian2013spatiotemporal,jain2014action,gkioxari2014finding,yu2015fast,weinzaepfel2015learning}.  
Analysis of broader surveillance scenes for action detection is also an active area of research.
Shu et al.~\cite{ShuXRTZ15} reason about groups of people, Loy et al.~\cite{LoyXG12} across multi-camera setups, and Kwak et al.~\cite{KwakHH13} based on quadratic programming-based instantiations of rules.  Common among these works is reasoning on spatio-temporal action proposals or human tracks, typically using sliding window-based approaches in the temporal dimension.  Furthermore, these works are in the context of trimmed or constrained-setting video clips.  
In contrast, we address the task of temporal action detection in untrimmed, unconstrained videos, with an efficient method for determining which frames to examine.

% Sermanet et al.~\cite{sermanet2013overfeat}, Szegedy et al.~\cite{szegedy2014scalable}, Erhan et al.~\cite{erhan2014scalable}, Girschick et al.~\cite{girshick2015fast}, Ren et al.~\cite{ren2015faster}, and Redmon et al.~\cite{redmon2015you}

\noindent \textbf{End-to-end detection.} Our goal of directly reasoning on the temporal bounds of actions shares philosophy with work in object detection that has regressed from full images to object bounds~\cite{sermanet2013overfeat, szegedy2014scalable, erhan2014scalable, girshick2015fast, ren2015faster, redmon2015you}. In contrast, existing action detection methods typically use exhaustive sliding-window approaches and post-processing to produce action instances~\cite{oneata2014lear, wangaction, karamanfast, yuanadsc}. To the best of our knowledge, our work is the first to address learning of temporal action detection in an end-to-end framework.

\noindent \textbf{Learning task-specific policies.} We draw inspiration from recent approaches that have used REINFORCE~\cite{williams1992simple} to learn task-specific policies. Mnih et al.~\cite{mnih2014recurrent}, Ba et al.~\cite{ba2014multiple}, and Sermanet et al.~\cite{sermanet2014attention} learn spatial attention policies for image classification, and Xu et al.~\cite{xu2015show} for image caption generation.  In a non-visual task, Zaremba et al.~\cite{zaremba2015reinforcement} learn policies for a Reinforcement Learning Neural Turing Machine.  Our method builds on these directions and uses REINFORCE to learn policies addressing the task of action detection.

\section{Method}

% \todo{This section needs a high-level opening.  Too much jargon in this paragraph.  The big picture is missing.}
% We introduce a single, unified framework for learning to predict the temporal extents of actions in video.  In order to learn in fully end-to-end fashion, we formulate our model as a reinforcement learning agent that decides both where to look and when to predict actions in a long video sequence.  We cast this in a recurrent neural network structure, and use a combination of backpropagation and the Reinforce algorithm~\cite{williams1992simple} to address training of the non-differentiable components of our model.

\begin{figure*}
\begin{center}
\includegraphics[width=0.95\linewidth]{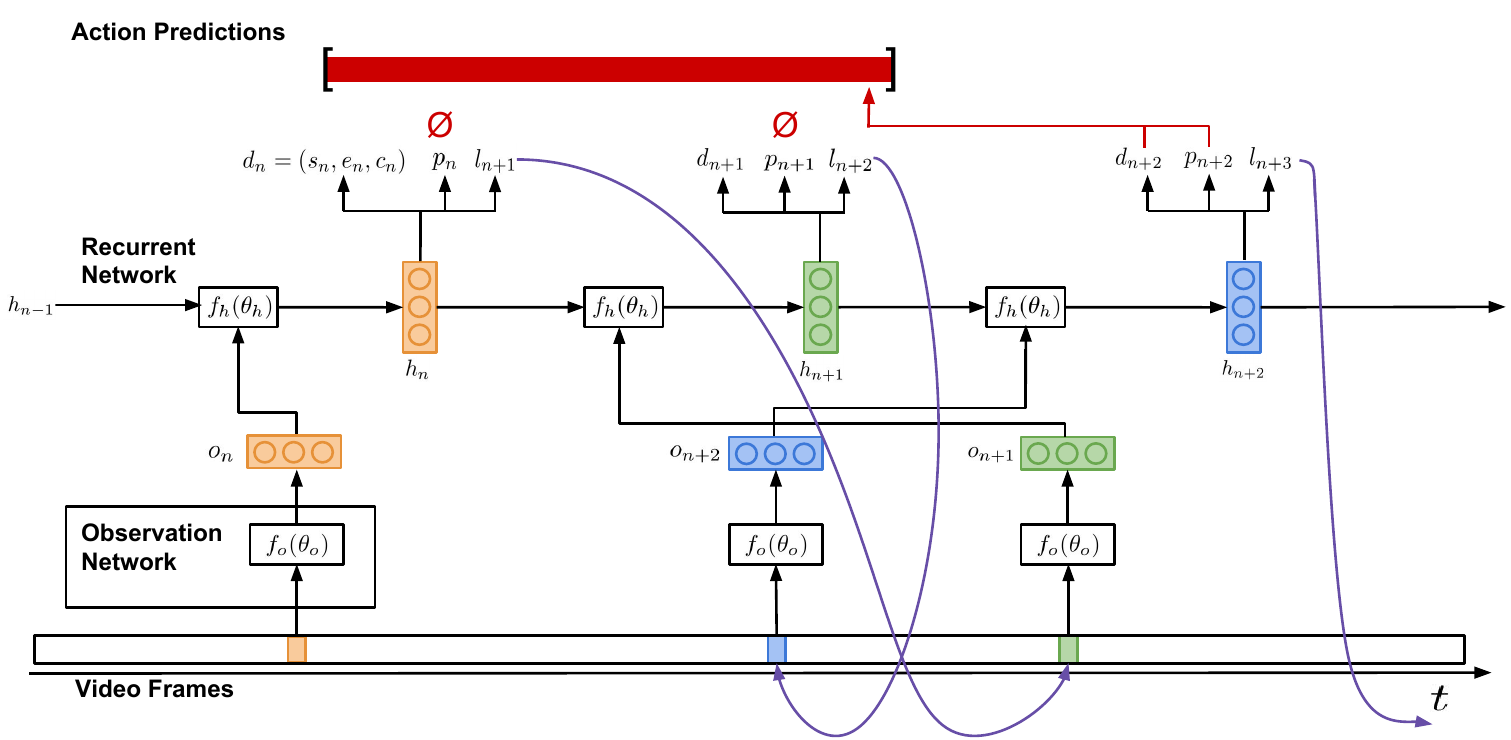}
\end{center}
\vspace{-0.1in}
  \caption{\small The input to the model is a sequence of video frames, and the output is a set of action predictions.  We illustrate an example of a forward pass. At timestep $n$, the agent observes the orange video frame and produces candidate detection $d_n$, however prediction indicator output $p_n$ suppresses it from being emitted into the prediction set. Observation location output $l_{n+1}$ signals to observe the the green video frame at the next timestep. The process repeats, and here again $p_{n+1}$ suppresses $d_{n+1}$ from being emitted. $l_{n+2}$ signals to now go backwards in the video to observe the blue frame. At timestep $n+2$, the action hypothesis is sufficiently refined, and the agent uses prediction indicator $p_{n+2}$ to emit $d_{n+2}$ into the prediction set (red arrow). The agent then continues proceeding through the video.}
\vspace{-0.1in}

\label{fig:model}
\end{figure*}

Our goal is to take a long sequence of video and output any instances of a given action. Fig.~\ref{fig:model} shows our model structure. The model is formulated as a reinforcement learning agent that interacts with a video over time.  The agent receives a sequence of video frames $\mathbf{V} = \{v_1,...,v_T\}$ as input, and can observe a fixed proportion of the frames. It must learn to effectively utilize these observations, or frame glimpses, to reason on the temporal bounds of actions.

% The goal here is to take a long sequence of videos and output both [blah blah and blah]. Fig.~\ref{fig:model} shows an overview of the model structure.  The agent is given a sequence of video frames $\mathbf{V} = \{v_1,...,v_T\}$ as input, and provided the ability to observe a fixed proportion of the frames. It must learn to effectively utilize these observations to reason on the temporal bounds of actions, and to output a potentially variable number of predicted action instances. \todo{not sure about agent vs model etc terminology}

\vspace{-0.02in}
\subsection{Architecture}
\label{sec:architecture}
% The model consists of two main components: an observation network (Sec.~\ref{sec:obs-net}), and a recurrent network (Sec.~\ref{sec:rec-net}). The \emph{observation network} encodes a visual representation of a video frame into observation vector $o_n$.  The \emph{recurrent network} sequentially processes video frames, using the observation network to process a video frame at each timestep, then updating its hidden state $h_n$ and deciding both whether to predict an action instance as well as which frame to observe next. We now describe each of them in more details. Later in Sec.~\ref{sec:training} we explain how we use a combination of backpropagation and the Reinforce algorithm to train the model in end-to-end fashion.

The model consists of two main components: an observation network (Sec.~\ref{sec:obs-net}), and a recurrent network (Sec.~\ref{sec:rec-net}). The \emph{observation network} encodes visual representations of video frames.  The \emph{recurrent network} sequentially processes these observations and decides both which frame to observe next and when to emit a prediction. We now describe each of these in more detail. Later in Sec.~\ref{sec:training}, we explain how we use a combination of backpropagation and REINFORCE to train the model in end-to-end fashion.

% At timestep $n$, the red video frame is observed and candidate detection $d_n$ is produced. The temporal extent of this candidate is shown for reference on the action predictions timeline, however the value of prediction indicator $p_n$ suppresses it from being emitted into the prediction set. This is indicated by the crossed out connection extending from $p_n$. Output $l_{n+1}$ indicates that the agent will observe the green video frame at the next timestep. The same process repeats, and once again $p_{n+1}$ suppresses candidate detection $d_{n+1}$ from being emitted. This time $l_{n+2}$ indicates that the agent will go backwards in the video to observe the blue frame. At timestep $n+2$, the agent has finally refined its action hypothesis sufficiently, and the red arrow indicates that it uses the value of prediction indicator $p_{n+2}$ to emit candidate detection $d_{n+2}$ into the prediction set.

\vspace{-0.1in}
\subsubsection{Observation Network} 
\label{sec:obs-net}

% [ALL VARIABLES NEED TO BE RIGOROUSLY DEFINED THROUGHOUT THE ENTIRE SEC 3. E.G. WHAT'S THETA?  WHAT'S THE RANGE OR DIMENSION OF O? ETC.] 

As shown in Fig.~\ref{fig:model}, the observation network $f_o$, parameterized by $\theta_o$, observes a single video frame at each timestep. It encodes the frame into a feature vector $o_n$ and provides this as input to the recurrent network.
 
Importantly, $o_n$ encodes information about both \textit{where} in the video an observation was taken as well as \textit{what} was seen. The inputs to the observation network therefore consist of the normalized temporal location of the observation, $l_n \in [0,1]$, and the corresponding video frame $v_{l_n}$.

% $l_n \in [0,1]$ is the normalized temporal location of an observation at time $n$.  Given a function $f_v$ mapping $l_n$ to its corresponding video frame index, the inputs to the observation network 

% The inputs to the observation network therefore consist of the normalized temporal location of the observation, $l_n \in [0,1]$, and the corresponding video frame $v_{f_v(l_n))}$. $f_v(l_n)$ is a function mapping $l_n$ to a video frame index.

% \todo{[REMIND YOUR READERS WHAT'S THE MAIN CHALLENGE HERE? ANYONE OFFERED A SOLUTION BEFORE? AND HOW TO MOTIVATE YOUR REPRESENTATION BELOW -- I.E. WHAT'S NOVEL AND NICE ABOUT THIS?]}

The architecture of the observation network is inspired by the spatial glimpse network of~\cite{mnih2014recurrent}.  Both $l_n$ and $v_{l_n}$ are mapped to a hidden space and then combined with a fully connected layer. $v_{l_n}$ is typically mapped with a sequence of convolutional, pooling, and fully connected layers; in our experiments we extract fc7 features from a fine-tuned VGG-16 network~\cite{simonyan2014very} and use $o_n \in \mathbb{R}^{1024}$.

\vspace{-0.12in}
\subsubsection{Recurrent Network}
\label{sec:rec-net}

The recurrent network $f_h$, parameterized by $\theta_h$, forms the core of the learning agent.  As can be seen in Fig.~\ref{fig:model}, the input to the network at each timestep $n$ is observation feature vector $o_n$. The network's hidden state $h_n$, a function of both $o_n$ and the previous hidden state $h_{n-1}$, models temporal hypotheses about action instances.

As the agent reasons on a video, three outputs are produced at each timestep: candidate detection $d_n$, binary indicator $p_n$ signaling whether to emit $d_n$ as a prediction, and temporal location $l_{n+1}$ indicating the frame to observe next.  We now describe each of these in more detail.

\vspace{0.1in}
\noindent \textbf{Candidate detection.} A candidate detection $d_n$ is produced using the function $d_n = f_d(h_n; \theta_d)$, where $f_d$ is a fully connected layer. $d_n$ is a tuple $(s_n, e_n, c_n) \in [0,1]^3$, where $s_n$ and $e_n$ are the normalized start and end locations of the detection, and $c_n$ is the confidence level of the detection. This candidate detection represents the agent's hypothesis surrounding a current action instance.  However, it is not emitted as a \textit{prediction} at each timestep, which would lead to a large amount of noise and many false positives. Instead, the agent uses a separate prediction indicator output to signal when a candidate detection should be emitted as a prediction. \\

\vspace{-0.1in}
\noindent \textbf{Prediction indicator.} The binary prediction indicator $p_n$  signals whether corresponding candidate detection $d_n$ should be emitted as a prediction. $p_n = f_p(h_n; \theta_p)$, where $f_p$ is a fully connected layer followed by a sigmoid nonlinearity. At training time, $f_p$ is used to parameterize a Bernoulli distribution from which $p_n$ is sampled; at test time, the maximum a posteriori estimate is used.

% Note that the decision of whether to output a candidate detection is not directly a function of the confidence level of the detection. Instead, it represents when the agent has acquired sufficient information about the action instance that it chooses to include the detection in its final set of predictions, while taking into account an objective to minimize false positives.  We explain how this policy is learned in Sec. 3.2.

The combination of the candidate detection and prediction indicator is crucial for the detection problem, where positive instances may occur anywhere or not at all. It enables the network to indicate when it has identified a unique action instance to add to the prediction set, and essentially folds non-maximum suppression in as a learnable component of our end-to-end framework.

% Having both the candidate detection and prediction indicator outputs is crucial for the detection setting, where instances may occur anywhere or not all, and gives the network the power to decide when it has identified a unique action instance that it wishes to add to the set of predicted actions. This essentially folds non-maximum suppression into our end-to-end framework.

\noindent \textbf{Location of next observation.} The temporal location $l_{n+1} \in [0,1]$ indicates the video frame that the agent chooses to observe next. This location is not constrained, and the agent may skip both forwards and backwards around a video.

The location is computed as $l_{n+1} = f_l(h_n; \theta_l)$, where $f_l$ is a fully connected layer, such that the agent's decision is a function of its past observations and their temporal locations. At training time, $l_{n+1}$ is sampled from a Gaussian distribution with a mean of $f_l(h_n; \theta_l)$ and a fixed variance; at test time, the maximum a posteriori estimate is used.

Fig.~\ref{fig:model} further illustrates the roles of these outputs and their interaction with an example of a forward pass through the network.

% \subsubsection{Forward Pass}
% \label{sec:forward_pass}
%  We illustrate a full forward pass of the network using Fig.~\ref{fig:model}. At timestep $n$, the red video frame is observed and candidate detection $d_n$ is produced. The temporal extent of this candidate is shown for reference on the action predictions timeline, however the value of prediction indicator $p_n$ suppresses it from being emitted into the prediction set. This is indicated by the crossed out connection extending from $p_n$. Output $l_{n+1}$ indicates that the agent will observe the green video frame at the next timestep. The same process repeats, and once again $p_{n+1}$ suppresses candidate detection $d_{n+1}$ from being emitted. This time $l_{n+2}$ indicates that the agent will go backwards in the video to observe the blue frame. At timestep $n+2$, the agent has finally refined its action hypothesis sufficiently, and the red arrow indicates that it uses the value of prediction indicator $p_{n+2}$ to emit candidate detection $d_{n+2}$ into the prediction set.

\subsection{Training}
\label{sec:training}

Our end goal is to learn to output a set of detected actions. To achieve this, we need to train the three outputs at each step of the agent's recurrent network: candidate detection $d_n$, prediction indicator $p_n$, and next observation location $l_{n+1}$. Given supervision from temporal action annotations in long videos, training these involves challenges of designing suitable loss and reward functions, and handling non-differentiable model components. We now explain how we address these challenges. We use standard backpropagation to train $d_n$, and REINFORCE to train $p_n$ and $l_{n+1}$.

\vspace{-0.1in}
\subsubsection{Candidate detections}

Candidate detections are trained using backpropagation to maximize the correctness of each candidate.  We wish to maximize correctness regardless of whether a candidate is ultimately emitted, since the candidates encode the agent's hypotheses about actions. This requires matching each candidate with a ground truth instance during training. We use the insight that at each timestep, the agent should form a hypothesis around the action instance (if any) nearest its current location in the video. This enables us to design a simple yet effective matching function.

\vspace{0.1in}
\noindent \textbf{Matching to ground truth.}  Given a set of candidate detections $D=\{d_n | n=1,...,N\}$ produced by a recurrent network of $N$ timesteps, and given ground truth action instances $g_{1,...,M}$, each candidate is matched to one ground truth instance, or none if $M = 0$. 

We define matching function
\vspace{-0.02in}
\begin{equation}
y_{nm} = \begin{cases}
1 &\text{if $m = \arg \min_{j=1,...,M} dist(l_n, g_j)$}\\
0 &\text{otherwise}
\end{cases}
\end{equation}
\vspace{-0.1in}

In other words, candidate $d_n$ is matched to ground truth $g_m$ if the agent's temporal location $l_n$ at timestep $n$ is closer to $g_m$ than any other ground truth instance. Defining $g_m = (s_m, e_m)$ as the start and end location of a ground truth instance, $dist(l_n, g_m)$ is simply $\min(|s_m-l_n|, |e_m-l_n|)$.  

% If $M = 0$ and the video sequence does not contain any ground truth instances, then $d_n$ is not matched to any ground truth and $y_{nm} = 0$ for all $m$.  We therefore have $\sum_i y_{nm} = \mathbbm{1}[M > 0]$. \\

\vspace{0.1in}
\noindent \textbf {Loss function.} Once candidate detections have been matched to ground truth instances, we optimize a multi-task classification and localization loss function over the set $D$:

\vspace{-0.12in}
\small
\begin{align}
L(D) = \sum_n L_{cls}(d_n) + \gamma \sum_n \sum_m \mathbbm{1}[y_{nm} = 1] L_{loc}(d_n, g_m)
\end{align}
\normalsize
\vspace{-0.12in}

Here the classification term $L_{cls}(d_n)$ is a standard cross-entropy loss on the detection confidence $c_n$, encouraging the confidence to be closer to 1 if detection $d_n$ is matched to a ground truth instance, and 0 otherwise.

If the detection is matched to a ground truth $g_m$ (i.e. $y_{nm} = 1$), the localization term $L_{loc}(d_n, g_m)$ is an $L_2$-regression loss that further encourages minimizing the distance $\lVert (s_n, e_n) - (s_m, e_m)\rVert$ between the two segments.

We optimize this loss function using backpropagation.

\vspace{-0.09in}
\subsubsection{Observation and emission sequences}

The observation location and prediction indicator outputs are non-differentiable components of our model that cannnot be trained with standard backpropagation. However, REINFORCE~\cite{williams1992simple} is a powerful approach that enables learning in non-differentiable settings. We first briefly describe REINFORCE below. We then introduce a reward function that we use with REINFORCE to learn effective policies for observation and prediction emission sequences.

% The non-differentiable prediction indicator and location outputs cannot be trained with standard backpropagation.  However, the Reinforce algorithm~\cite{williams1992simple} enables learning with non-differentiable operators. We use it here to learn policies for the prediction indicator and location output sequences. \\
\vspace{0.1in}
\noindent \textbf{REINFORCE.} Given $\mathcal{A}$, a space of action sequences, and $p_\theta(a)$, a distribution over $a \in \mathcal{A}$ and parameterized by $\theta$, the REINFORCE objective can be expressed as

\vspace{-0.15in}
\small
\begin{align}
J(\theta) = \sum_{a \in \mathcal{A}} p_\theta(a)r(a)
\end{align}
\normalsize
\vspace{-0.1in}

Here $r(a)$ is a reward assigned to each possible action sequence, and $J(\theta)$ is the expected reward under the distribution of possible action sequences. In our case we wish to learn network parameters $\theta$ that maximize the expected reward of a sequence of location and prediction indicator outputs.

The gradient of the objective is

\vspace{-0.15in}
\small
\begin{align}
\nabla J(\theta) &= \sum_{a \in \mathcal{A}} p_\theta(a) \nabla \log p_\theta(a) r(a)
\end{align}
\normalsize
\vspace{-0.1in}

This leads to a non-trivial optimization problem due to the high-dimensional space of possible action sequences. REINFORCE addresses this by learning network parameters using Monte Carlo sampling and an approximation to the gradient equation:

\vspace{-0.15in}
\small
\begin{align}
\nabla J(\theta) \approx \frac{1}{K} \sum_{i=1}^K \sum_{n=1}^N \nabla \log \pi_\theta(a_n^i | h_{1:n}^i, a_{1:n-1}^i) R_n^i
\end{align}
\normalsize
\vspace{-0.05in}

\begin{table*}[t]
% \footnotesize
\small
\begin{center}
\begin{tabularx}{.75\textwidth}{
p{.25\textwidth}X<{\hfill}
>{\centering}p{.1\textwidth}X
>{\centering}p{.1\textwidth}X
>{\centering}p{.1\textwidth}X
>{\centering}p{.1\textwidth}X
>{\centering}p{.1\textwidth}X
}
\hline
& $\alpha$=0.5 & $\alpha$=0.4 & $\alpha$=0.3 & $\alpha$=0.2 & $\alpha$=0.1\\
\hline
Karaman et al.~\cite{karamanfast} & 0.9 & 1.4 & 2.1 & 3.4 & 4.6\\
Wang et al.~\cite{wangaction} &  8.3 & 11.7 & 14.0 & 17.0 & 18.2\\
Oneata et al.~\cite{oneata2014lear} &  14.4 & 20.8 & 27.0 & 33.6 & 36.6\\
\hline
\textbf{Ours (full)} & \textbf{17.1} & \textbf{26.4} &  \textbf{36.0} & \textbf{44.0} & \textbf{48.9}\\
\hline
\hline
\multicolumn{6}{c}{Ablation Experiments}\\
\hline
\hline
Ours w/o $d_{pred}$ & 12.4 & 19.3 & 26.0 & 32.5 & 37.0\\
Ours w/o $d_{obs}$ & 9.3 & 15.2 & 20.6 & 26.5 & 31.2\\
Ours w/o $d_{obs}$ w/o $d_{pred}$ & 8.6 & 14.6 & 20.0 & 27.1 & 33.3\\
% LSTM with NMS & - & - & - & - & -\\
Ours w/o $loc$ & 5.5 & 9.9 & 16.2 & 22.7 & 27.5\\
CNN with NMS & 6.4 & 9.6 & 12.8 & 16.7 & 18.5\\
LSTM with NMS & 5.6 & 7.8 & 10.3 & 13.9 & 15.7\\
\hline
\end{tabularx}
\end{center}
\vspace{-0.2in}
\caption{\small \small Action detection results on THUMOS'14. Comparison with the top 3 performers on the THUMOS'14 challenge leaderboard is shown, as well as with ablation models. mAP is reported for different intersection-over-union (IOU) thresholds $\alpha$.}
\label{table:thumos_mAP}
\end{table*}

Given an agent interacting with an environment, in our case a video, $\pi_\theta$ is the agent's policy. This is a learned distribution over actions conditioned on the interaction sequence thus far. At timestep $n$, $a_n$ is the policy's current action (e.g. location $l_{n+1}$ or prediction indicator $p_n$), $h_{1:n}$ is the history of past states including the current, and $a_{1:n-1}$ is the history of past actions. $R_n = \sum_{t=n}^N r_t$ is the cumulative future reward obtained from the current timestep onward, for a sequence of $N$ timesteps. The approximate gradient is computed by running an agent's current policy in its environment to obtain $K$ interaction sequences.

To reduce the variance of the gradient estimate, a baseline reward $b_n^i$ is often estimated, e.g. via a separate network, and subtracted so that the gradient equation becomes:

\vspace{-0.15in}
\small
\begin{align}
\nabla J(\theta) \approx \frac{1}{K} \sum_{i=1}^K \sum_{n=1}^N \nabla \log \pi_\theta(a_n^i | h_{1:n}^i, a_{1:n-1}^i) (R_n^i - b_n^i)
\end{align}
\normalsize
\vspace{-0.15in}

REINFORCE learns model parameters according to this approximate gradient. The log-probability $\log \pi_\theta$ of actions leading to high future reward are increased, and those leading to low reward are decreased. Model parameters can then be updated using backpropagation. \\

\noindent \textbf{Reward function.}  Training with REINFORCE requires designing an appropriate reward function. Our goal is to learn policies for the location and prediction indicator outputs that lead to action detection with both high recall and high precision. We therefore introduce a reward function that seeks to maximize true positive detections while minimizing false positives:

\vspace{-0.1in}
\small
\begin{equation}
r_N = \begin{cases}
R_{p} &\text{if $M > 0$ and $N_p = 0$}\\
N_{+}R_{+} + N_{-}R_{-} &\text{otherwise}
\end{cases}
\end{equation}
\normalsize
\vspace{-0.1in}

All reward is provided at the $N$th (final) timestep, and is 0 for $n < N$, since we want to learn policies that jointly lead to high overall detection performance. $M$ is the number of ground truth action instances, and $N_p$ is the number of predictions emitted by the agent. $N_{+}$ is the number of true positive predictions, $N_{-}$ is the number of false positive predictions, and $R_{+}$ and $R_{-}$ are positive and negative rewards contributed by each of these predictions, respectively. A prediction is considered correct if its overlap with a ground truth is both greater than a threshold and higher than that of any other prediction. In order to encourage the agent not to be overly conservative, a negative reward $R_p$ is provided if the the video contains ground truth instances ($M > 0$) but the model did not emit any predictions ($N_p = 0$).

We use this function with REINFORCE to train the location and prediction indicator outputs, and learn observation and emission policies optimized for action detection.

% train these outputs with a reward function that seeks to maximize true positive detections while minimizing false positives.  Specifically, the reward function is defined as

\section{Experiments}

We evaluate our model on two datasets - THUMOS'14~\cite{THUMOS14} and ActivityNet~\cite{caba2015activitynet}.  We show that our end-to-end approach enables the model to outperform state-of-the-art results by a large margin on both datasets.  Furthermore, the learned policy of frame observations is both effective and efficient; the model achieves these results while observing in total only 2\% or less of the video frames.

\subsection{Implementation Details}
\label{sec:implementation}
We learn a 1-vs-all model for each action class. In the observation network, we use a VGG-16 network~\cite{simonyan2014very} fine-tuned on the dataset to extract visual features from observed video frames. Fc7-layer features are extracted and embedded with the temporal location of the frame into a 1024-dimensional observation vector.

For the recurrent network, we use a 3-layer LSTM network with 1024 hidden units in each layer. Videos are downsampled to 5fps in THUMOS'14 and 1fps in ActivityNet, and processed in sequences of 50 frames. The agent is given a fixed number of observations for each sequence, typically 6 in our experiments. All temporal locations are normalized to $[0,1]$ in a video sequence. Any predictions overlapping or crossing sequence bounds are merged with a simple union rule. We learn with mini-batches of 256 sequences, and use RMSProp~\cite{rmsprop} to modulate the per-parameter learning rate during optimization.  Other hyperparameters are learned through cross-validation. The ratio of sequences containing positive examples in each mini-batch is an important hyperparameter to prevent the model from being overly conservative. Approximately one-third to one-half positive examples is typically used.

\subsection{THUMOS'14 Dataset}
\label{sec:thumos}
% \subsubsection{Dataset}

% The action detection task of THUMOS'14 consists of 20 classes of sports.   Training data consists of temporal annotations in 1010 untrimmed videos, as well as 13,320 trimmed clips and 2500 background videos with no positive instances. The test set consists of 1500 untrimmed videos.  Since the detection task comprises only 20 of the 101 action classes in the dataset, we first coarsely filter these videos for each class, using video-level average pooling over class probabilties computed every 300 frames, or .003 fps.

% Evaluation is performed using mean average precision (mAP). Predictions are considered correct if their intersection-over-union (IOU) with a ground truth instance is greater than a threshold $\alpha$, and the mAP over all classes is computed.

% Each video contains on average 5 action instances, and actions have an average length of 2.3 seconds.
% \subsubsection{Results}

The action detection task of THUMOS'14~\cite{THUMOS14} consists of 20 classes of sports, and Table~\ref{table:thumos_mAP} shows results on this dataset. Since the task comprises only 20 of the 101 action classes in the dataset, we first coarsely filter the full set of test videos for these classes, using video-level average pooling over class probabilities that are computed every 300 frames (0.1 fps).  We report mAP for different IOU thresholds $\alpha$, and compare with the top 3 performers on the THUMOS'14 challenge leaderboard~\cite{THUMOS14}. All these methods compute dense trajectories~\cite{wang2013action} and/or CNN features over temporal windows, and use a sliding window approach with non-maximum suppression to obtain predictions. ~\cite{karamanfast} uses dense trajectories only,~\cite{wangaction} uses temporal windows of combined dense trajectories and CNN features, and~\cite{oneata2014lear} uses temporal windows of dense trajectories with video-level CNN classification predictions. 

Our model outperforms existing methods at all values of $\alpha$. The relative margin increases as we decrease $\alpha$, indicating that our model more frequently predicts actions near ground truth instances even when not precisely localized. Our model achieves these results while processing only 2\% of videos frames using its learned observation policy.

% The action detection task of THUMOS'14 consists of 20 classes of sports. Since the task comprises only 20 of the 101 action classes in the dataset, we first coarsely filter test videos for each class, using video-level average pooling over class probabilties that computed every 300 frames (.003 fps). Evaluation is performed using mean average precision (mAP) over all classes, where a prediction is considered correct if its intersection-over-union (IOU) with a ground truth instance is greater than a threshold $\alpha$.

% Table~\ref{table:thumos_mAP} shows results on this dataset. We report mAP for different IOU thresholds $\alpha$, and compare with the top 3 performers on the THUMOS'14 challenge leaderboard~\cite{THUMOS14}. All these methods compute dense trajectories~\cite{wang2013action} and/or CNN features over temporal windows, and use a sliding window approach with non-maximum suppression to obtain predictions. ~\cite{karamanfast} uses dense trajectories only,~\cite{wangaction} uses temporal windows of combined dense trajectories and CNN features, and~\cite{oneata2014lear} uses temporal windows of dense trajectories with video-level CNN classification predictions. 

% \todo{Ours -> an acronym?  Not sure, there's lots of "Ours" floating around.}

\begin{figure}[t]
\begin{center}
\includegraphics[width=\linewidth]{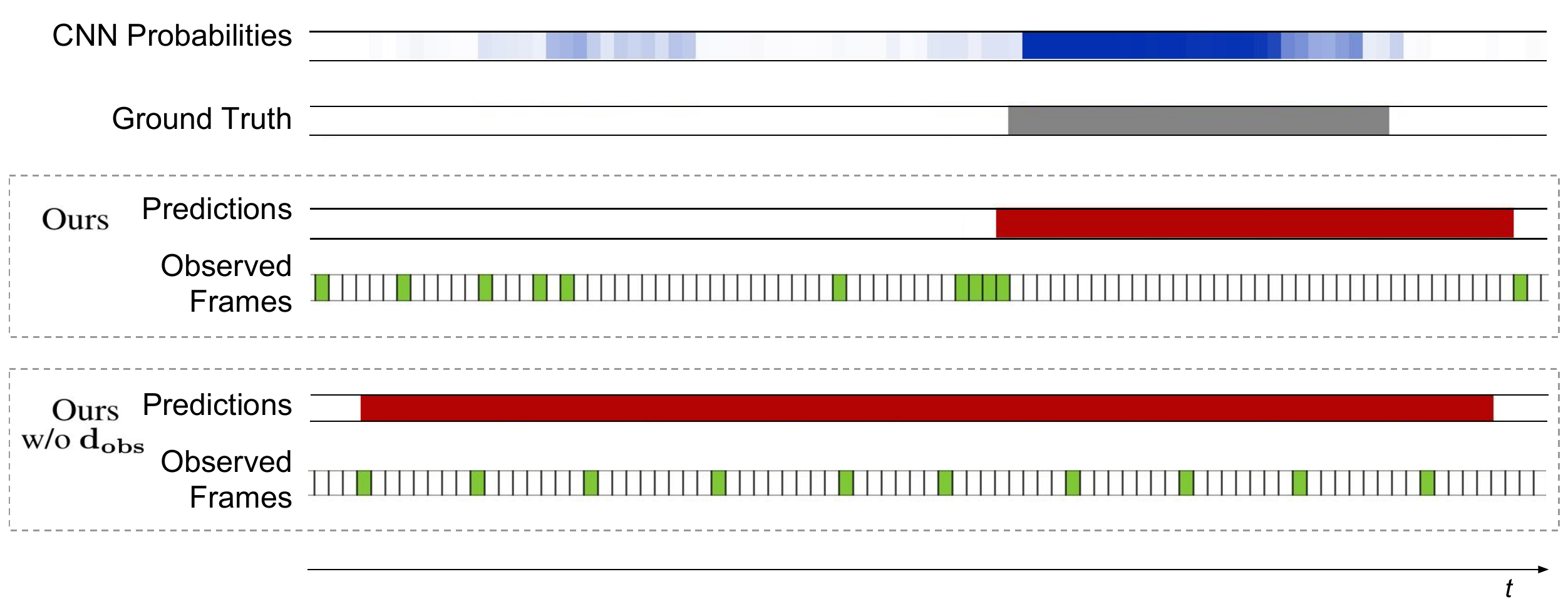}
\end{center}
\vspace{-0.25in}
  \caption{\small Comparison of Our full model with the Ours w/o $d_{obs}$ model. Refer to Fig.~\ref{fig:thumos-qualitative} caption for explanation of figure structure and color scheme. Each model's observed frames are shown in green, and the prediction extent in red. Allowing the model to choose which frames to observe enables the necessary resolution to reason precisely on action bounds.}
\label{fig:long}
\label{fig:ablation}
\end{figure}

\noindent \textbf{Ablation experiments.} Table~\ref{table:thumos_mAP} also shows results for ablation experiments analyzing the contributions of different model components. The ablation models are as follows:
\begin{itemize}
\vspace{-0.1in}
\item \textbf{Ours w/o $\mathbf{d_{pred}}$} removes the prediction indicator output. The candidate detection at every timestep is emitted, and merged with non-maximum suppression.
\vspace{-0.1in}
\item \textbf{Ours w/o $\mathbf{d_{obs}}$} removes the location output indicating where to observe next. Observations are instead determined by uniform sampling with the same total number of observations.
\vspace{-0.1in}
\item \textbf{Ours w/o $\mathbf{d_{obs}}$ w/o $\mathbf{d_{pred}}$} removes both the prediction indicator and location observation outputs.
\vspace{-0.1in}
\item \textbf{Ours w/o $\mathbf{loc}$} removes localization regression. All emitted detections are of median length from the training set, and centered on the currently observed frame.
% \item \textbf{LSTM with NMS} removes direct prediction of action instance bounds. Instead per-frame class probabiltiies are obtained and aggregated using temporal windows of multiple lengths and non-maximum suppression (NMS), similar to previous work.
\vspace{-0.25in}
\item \textbf{CNN with NMS} removes direct prediction of temporal action bounds. Per-frame class probabilities from the VGG-16 Network~\cite{simonyan2014very} used in our observation network are densely obtained at multiple temporal scales and aggregated with non-maximum suppression, similar to existing work~\cite{karamanfast, wangaction, oneata2014lear}.
\vspace{-0.2in}
\begin{table}[t]
\footnotesize
\centering
\begin{tabulary}{\linewidth}{|L|C|C||L|C|C|}
\hline
&~\cite{oneata2014lear} & Ours & &~\cite{oneata2014lear} & Ours\\
\hline
\hline
Baseball Pitch & 8.6 & \textbf{14.6} & Hamm. Throw & \textbf{34.7} & 28.9\\
Basket. Dunk & 1.0 & \textbf{6.3} & High Jump & 17.6 & \textbf{33.3}\\
Billiards & 2.6 & \textbf{9.4} & Javelin Throw & \textbf{22.0} & 20.4\\
Clean and Jerk & 13.3 & \textbf{42.8} & Long Jump & \textbf{47.6} & 39.0\\
Cliff Diving & \textbf{17.7} & 15.6 & Pole Vault & \textbf{19.6} & 16.3\\
Cricket Bowl. & 9.5 & \textbf{10.8} & Shotput & 11.9 & \textbf{16.6}\\
Cricket Shot & 2.6 & \textbf{3.5} & Soccer Penalty & \textbf{8.7} & 8.3\\
Diving & 4.6 & \textbf{10.8} & Tennis Swing & 3.0 & \textbf{5.6}\\
Frisbee Catch & 1.2 & \textbf{10.4} & Throw Discus & \textbf{36.2} & 29.5\\
Golf Swing & \textbf{22.6} & 13.8 & Volley. Spike & 1.4 & \textbf{5.2}\\
\hline
\hline
\multicolumn{3}{|l}{\textbf{mAP}} &   & 14.4 & \textbf{17.1}\\
\hline
\end{tabulary}
\vspace{-0.05in}
\caption{\small Per-class breakdown (AP) on THUMOS'14, at IOU of $\alpha=0.5$.}
\label{table:thumos_perclass}
\end{table}
\vspace{0.2in}

\end{itemize}

\begin{figure}[t]
\begin{center}
\includegraphics[width=\linewidth]{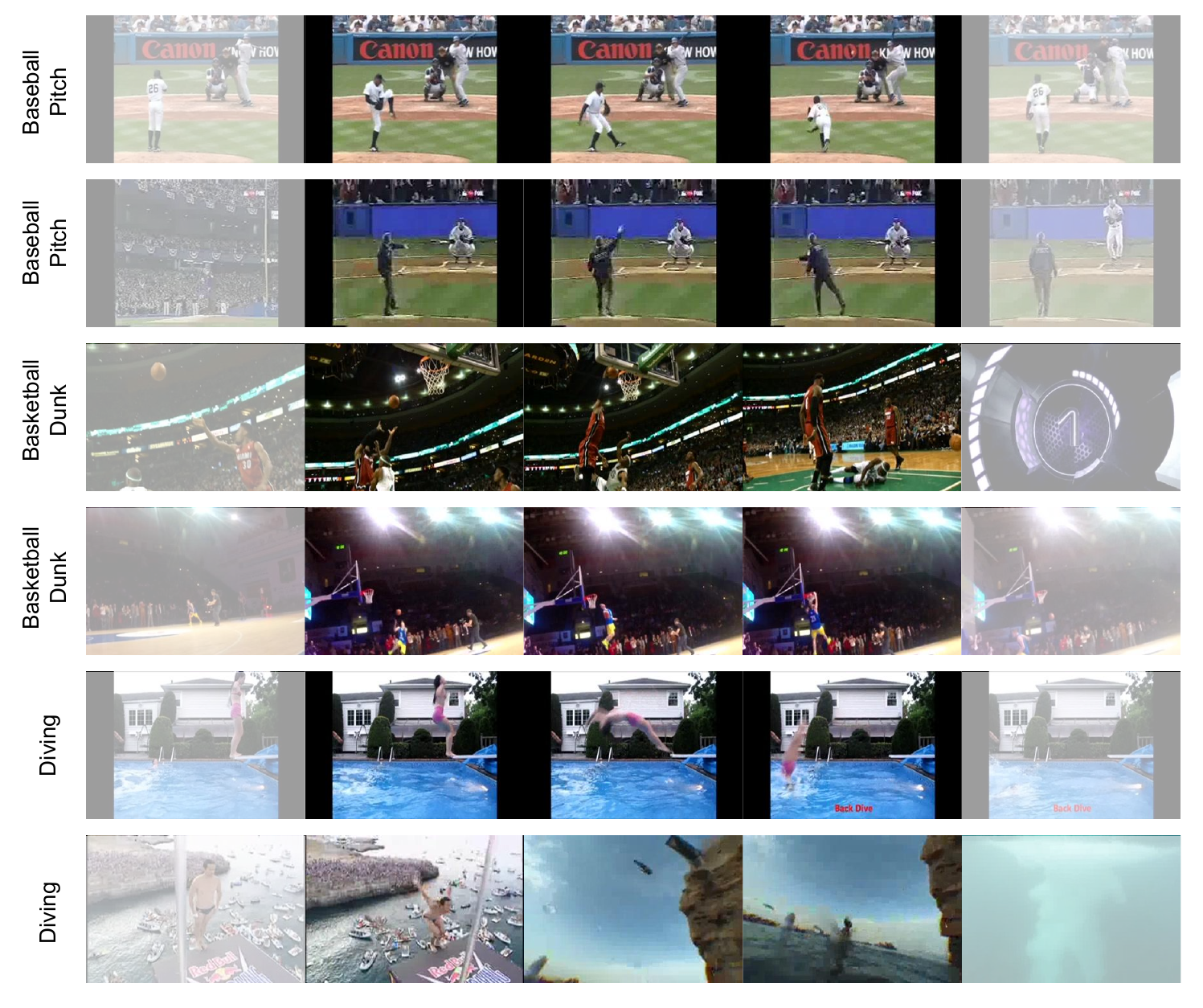}
\end{center}
\vspace{-0.25in}
  \caption{\small Examples of predicted action instances on THUMOS'14. Each row shows sampled frames during or just outside the temporal extent of a detected action. Faded frames indicate location outside the detection and illustrate localization ability.}
\label{fig:long}
\label{fig:det_examples}
\end{figure}

\begin{figure*}[t]
    \centering
    \begin{subfigure}{0.9\linewidth}
    \includegraphics[width=\linewidth]{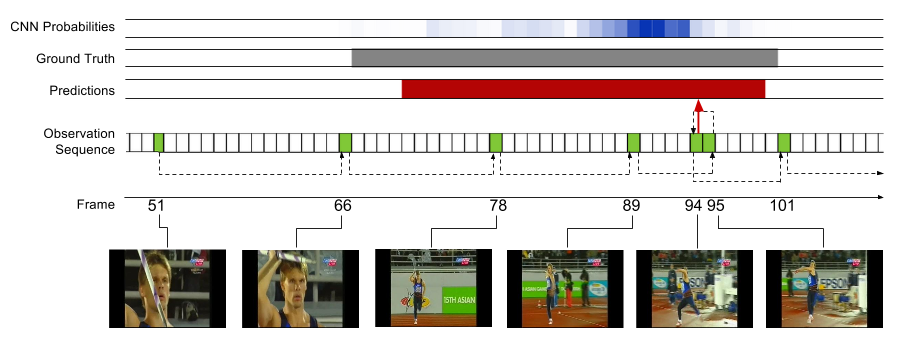}
    \end{subfigure}
    \begin{subfigure}{0.9\linewidth}
    \includegraphics[width=\linewidth]{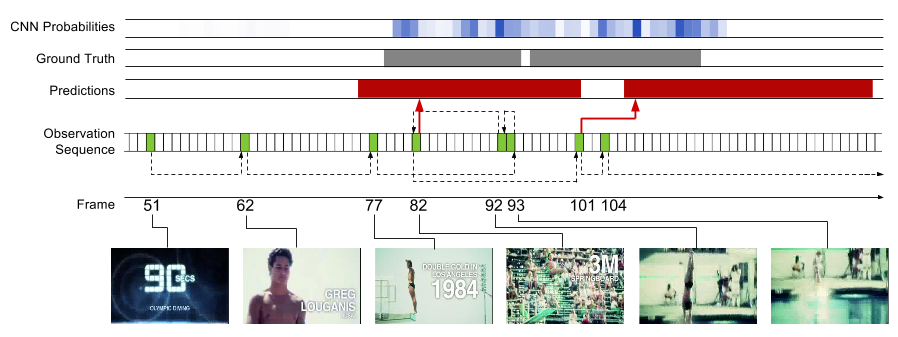}
    \end{subfigure}
    \vspace{-0.1in}
    \caption{\small Examples of learned observation policies on THUMOS'14. The top example shows a javelin throw and the bottom example shows diving. Observed frames are colored in green and labeled with the frame index. Prediction extents are shown in red, and ground truth in grey. For reference, we also show frame-level CNN probabilities from the VGGNet used in our observation network; higher intensity indicates higher probability and provides insight into frame-level signal for the class. Dashed arrows indicate the observation sequence, and red arrows indicate frames where a prediction was emitted.}
    \label{fig:thumos-qualitative}
\end{figure*}

\vspace{-0.05in}
Ours w/o $d_{pred}$ obtains lower performance compared to the full model, due to many false positives. Ours w/o $d_{obs}$ also lowers performance since uniform sampling does not provide sufficient resolution to localize action boundaries (Fig. \ref{fig:ablation}). Interestingly, removing $d_{obs}$ cripples the model more than removing $d_{pred}$, highlighting the importance of the observation policy. As expected, removing both outputs in Ours w/o $d_{obs}$ w/o $d_{pred}$ decreases performance further. Ours w/o $loc$ is the poorest performing model at $\alpha=0.5$, even below the CNN, showing the importance of temporal regression. The relative difference with the CNN decreases and then flips when we decrease $\alpha$, indicating that the model still detects the rough location of actions but suffers from imprecise localization. Finally, the CNN with NMS achieves significantly lower performance than all ablation models except the Ours w/o $loc$ model, quantifying the contribution of our end-to-end framework. Its performance is also in the range of but lower than \cite{wangaction}, which uses dense trajectories~\cite{wang2013action} and Imagenet-pretrained~\cite{russakovsky2014imagenet} CNN features. This suggests that additionally incorporating motion-based features would further improve the performance of our model.

As an additional baseline, we perform NMS on top of an LSTM, a standard
temporal network which produces frame-level smoothing and consistency~\cite{donahue2014long}. The LSTM with NMS achieves lower
performance than the CNN with NMS, despite adding greater temporal
consistency. The main reason appears to be that increasing the
temporal smoothness of frame-level class probabilities is actually
harmful, not beneficial, to the task of action instance detection,
where precise localization of temporal boundaries is required.

Finally, we experimented with different numbers of observations per video sequence, e.g. 4, 8, and 10.  Detection performance was not substantially different across this range. This is consistent with other work on CNNs for action recognition using max-pooling ~\cite{zha2015exploiting}, highlighting the importance of learning effective frame observation policies.

% We additionally experimented with different numbers of observations per video sequence, for example 4, 8, and 10, which did not lead to consistent trends in detection performance. This supports insight from other work that judicious compression of frames can greatly mitigate performance degredation~\cite{zha2015exploiting}, and highlights the importance of learning effective frame observation policies.

\noindent\textbf{Per-class breakdown.}
Table~\ref{table:thumos_perclass} shows the per-class AP breakdown of our model, and comparison with the top performer~\cite{oneata2014lear} on the THUMOS'14 leaderboard. Our model outperforms~\cite{oneata2014lear} on 12 out of 20 classes. Notably, it shows significant improvement on some of the most challenging classes in the dataset such as basketball dunk, diving, and frisbee catch. Fig.~\ref{fig:det_examples} shows examples of our model's predictions, including several from these challenging classes. The model's ability to reason holistically on action extents enables it to infer temporal boundaries even when frame appearance is challenging: e.g. similar pose and environment, or abrupt scene change in the second diving example.

% The final two experiments, LSTM with NMS and CNN with NMS, show the representation power of the raw frame observation features that our input into our model. The CNN with NMS experiment achieves slightly lower performance than~\cite{wangaction}, which includes dense trajectories.  This shows that the significant improvement from our model indeed comes from its various components and not the input features.  Furthermore, incorporating dense trajectory features into our model could likely further improve performance. 

\noindent \textbf{Observation policy analysis.} Fig.~\ref{fig:thumos-qualitative} shows examples of observation policies that our model learns, as well as accompanying predictions. For reference, we also show frame-level CNN probabilities from the VGGNet used in our observation network, to provide insight into frame-level signal for the action. In the top example of a javelin throw, the model begins to take more frequent observations once the person begins running. Near the end boundary of the action, it takes a step backwards to refine its hypothesis and then emits a prediction before moving on. 

\begin{figure*}[t]
    \centering
    \begin{subfigure}{0.9\linewidth}
    \includegraphics[width=\linewidth]{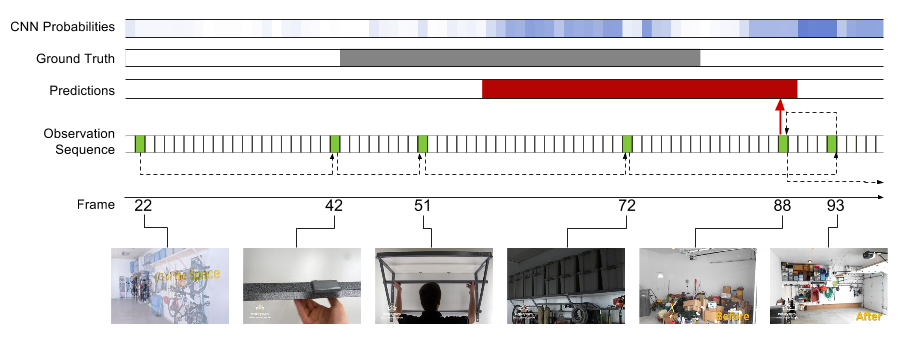}
    \end{subfigure}
    \vspace{-0.1in}
    \caption{\small Example of a learned observation policy on the Work subset of ActivityNet. The action is Organizing Boxes. Refer to Fig.~\ref{fig:thumos-qualitative} for explanation of figure structure and color scheme.}
    \label{fig:anet-qualitative}
\end{figure*}

The lower example of diving is a challenging case where two action instances occur in quick succession. While the strength of the frame-level CNN probabilities over the sequence would be difficult for standard sliding-window approaches to handle, our model is able to discern the two separate instances. The model once again takes steps backwards to refine its prediction, including once (frame 93) when motion blur makes it difficult to discern much from the frame. However, the predictions are also somewhat longer than the ground truth, and upon observing its first frame of the second instance (frame 101), the model immediately emits a prediction of comparable but slightly longer duration than the first.  This suggests that the model may have learned duration priors that, while generally beneficial, were overly strong in this case.

\subsection{ActivityNet Dataset}
The ActivityNet action detection dataset~\cite{caba2015activitynet} consists of 68.8 hours of temporal annotations in 849 hours of untrimmed, unconstrained video. There are 1.41 action instances per video and 193 instances per class.  Tables~\ref{table:anet_sports_perclass} and~\ref{table:anet_work_perclass} show per-class and mAP performance on the ``Playing sports'' and ``Work, main job'' subsets of ActivityNet, respectively. Evaluation uses the ActivityNet validation set, and hyperparameters are cross-validated on the training set.

Our model outperforms existing work~\cite{caba2015activitynet}, which is based on a combination of dense trajectories, SIFT, and ImageNet-pretrained CNN features, by significant margins. It outperforms \cite{caba2015activitynet} in 13 out of 21 classes on the Sports subset and in 10 out of 15 classes on the Work subset. The improvement is particularly large on the Work subset. This is partially attributable to the fact that work activities are generally less well-defined and have less discriminative movements. In the example sequence of the Organizing Boxes action in Fig.~\ref{fig:anet-qualitative}, this is evident in the weaker, more diffuse frame-level CNN probabilities for the action. While this creates a challenge for approaches that rely on post-processing, our model's direct reasoning on action extents enables it to still produce strong predictions.

\begin{table}[t]
\footnotesize
\centering
\begin{tabulary}{\linewidth}{|L|C|C||L|C|C|}
\hline
&~\cite{caba2015activitynet} & Ours & &~\cite{caba2015activitynet} & Ours\\
\hline
\hline
Archery & \textbf{34.7} & 5.2 & Long Jump & 41.1 & \textbf{56.8}\\
Bowling & 51.3 & \textbf{52.2} & Mountain Climb. & 31.0 & \textbf{53.0}\\
Bungee & 42.6 & \textbf{48.9} & Paintball & \textbf{31.2} & 12.5\\
Cricket & 27.9 & \textbf{38.4} & Playing Kickball & 33.8 & \textbf{60.8}\\
Curling & 16.4 & \textbf{30.1} & Playing Volley. & 32.1 & \textbf{40.2}\\
Discus Throw & \textbf{26.2} & 17.6 & Pole Vault & \textbf{47.7} & 35.5\\
Dodgeball & 26.6 & \textbf{61.3} & Shot put & 29.4 & \textbf{50.9}\\
Doing Moto. & 30.2 & \textbf{46.2} & Skateboarding & 21.3 & \textbf{34.4}\\
Ham. Throw & \textbf{22.2} & 13.7 & Start Fire & 25.3 & \textbf{38.4}\\
High Jump & \textbf{41.3} & 21.9 & Triple Jump & \textbf{36.4} & 16.1\\
Javelin Throw & \textbf{48.1} & 35.7 & & &\\
\hline
\hline
\multicolumn{3}{|l}{\textbf{mAP}} &   & 33.2 & \textbf{36.7}\\
\hline
\end{tabulary}
\caption{\small Per-class breakdown and mAP on the ActivityNet Sports subset, at IOU of $\alpha=0.5$.}
\vspace{-0.05in}
\label{table:anet_sports_perclass}
\end{table}

\begin{table}[t]
\footnotesize
\centering
\begin{tabulary}{\linewidth}{|L|C|C||L|C|C|}
\hline
&~\cite{caba2015activitynet} & Ours & &~\cite{caba2015activitynet} & Ours\\
\hline
\hline
Attend Conf. & 28.3 & \textbf{56.5} & Phoning & 34.7 & \textbf{52.1}\\
Search Security & 24.5 & \textbf{33.9} & Pumping Gas & \textbf{54.7} & 34.0\\
Buy Fast Food & 34.4 & \textbf{45.8} & Setup Comp. & \textbf{37.4} & 30.3\\
Clean Laptop Fan & 26.0 & \textbf{35.8} & Sharp. Knife & \textbf{36.3} & 35.2\\
Making Copies & 18.2 & \textbf{41.7} & Sort Books & \textbf{29.3} & 16.7\\
Organizing Boxes & \textbf{29.6} & 19.1 & Using Comp. & 37.4 & \textbf{50.2}\\
Organiz. Cabin. & 19.0 & \textbf{43.7} & Using ATM & 29.5 & \textbf{64.9}\\
Packing & 28.0 & \textbf{39.1} & & &\\
\hline
\hline
\multicolumn{3}{|l}{\textbf{mAP}} &   & 31.1 & \textbf{39.9}\\
\hline
\end{tabulary}
\caption{\small Per-class breakdown and mAP on the ActivityNet Work subset, at IOU of $\alpha=0.5$.}
\vspace{-0.05in}
\label{table:anet_work_perclass}
\end{table}

% \vspace{-0.05in}
\section{Conclusion}
% \vspace{-0.05in}
In conclusion, we have introduced an end-to-end approach for action detection in videos that learns to directly predict the temporal bounds of actions. Our model achieves state-of-the-art results on the THUMOS'14 and ActivityNet action detection datasets while observing only a fraction of frames. A direction for future work is to extend our framework to learn joint spatio-temporal observation policies.

% \vspace{-0.1in}
\section{Acknowledgments}
% \vspace{-0.1in}
We would like to thank Juan Carlos Niebles and Fabian Caba Heilbron for help with ActivityNet baselines.

{\small
\bibliographystyle{ieee}
\bibliography{egbib}
}

\end{document}